# Fatigue-Aware Adaptive Interfaces for Wearable Devices Using Deep Learning

YIKAN WANG *[a]

[a]Stevens Institute of Technology, NJ 07030, USA

*[a]ywang463@stevens.edu

**ABSTRACT**

Wearable devices, such as smartwatches and head-mounted displays, are increasingly used for prolonged tasks like remote learning and work, but sustained interaction often leads to user fatigue, reducing efficiency and engagement. This study proposes a fatigue-aware adaptive interface system for wearable devices that leverages deep learning to analyze physiological data (e.g., heart rate, eye movement) and dynamically adjust interface elements to mitigate cognitive load. The system employs multimodal learning to process physiological and contextual inputs and reinforcement learning to optimize interface features like text size, notification frequency, and visual contrast. Experimental results show a 18% reduction in cognitive load and a 22% improvement in user satisfaction compared to static interfaces, particularly for users engaged in prolonged tasks. This approach enhances accessibility and usability in wearable computing environments.

**Keywords:** Deep learning, Adaptive interfaces, Wearable devices, Fatigue detection, Human-computer interaction, Accessibility

## 1. INTRODUCTION

Wearable devices have transformed human-computer interaction (HCI) by enabling continuous, context-aware computing in scenarios such as remote learning, telework, and health monitoring. However, prolonged interaction with small screens and high-frequency notifications can induce cognitive and physical fatigue, particularly for users with cognitive or visual impairments [1]. Static interface designs fail to address dynamic changes in user state, leading to reduced task performance and disengagement.

Recent advances in deep learning, particularly in physiological signal processing and reinforcement learning, offer opportunities to create adaptive interfaces that respond to user fatigue in real time [2]. By analyzing data from wearable sensors (e.g., heart rate, eye movement, galvanic skin response), systems can infer fatigue levels and adjust interface elements to optimize user experience. This study proposes a fatigue-aware adaptive interface system for wearable devices, integrating multimodal learning to process physiological and contextual data and reinforcement learning to dynamically adjust interface parameters. The system aims to enhance accessibility and usability, especially for users engaged in prolonged tasks.

This work makes three key contributions to the HCI field:

1. A novel fatigue-aware adaptive interface framework that integrates multimodal deep learning to accurately detect user fatigue states from physiological and contextual inputs, enhancing accessibility for diverse user groups.

2. A reinforcement learning-based optimization strategy that dynamically adjusts interface elements (e.g., text size, notification frequency) to minimize cognitive load during prolonged tasks.

3. An empirical evaluation demonstrating significant improvements in user satisfaction and task efficiency, particularly for users with visual or cognitive impairments, validated through real-world wearable device scenarios.

## 2. RELATED WORK

The development of adaptive interfaces in HCI has been a focal point for improving accessibility and user experience across various platforms. Early work by Krimmer et al. [3] proposed interoperability frameworks for e-government systems, emphasizing dynamic adjustments to accommodate diverse user needs, such as those with low literacy or disabilities. In the context of wearable devices, Qu et al. [4] explored physiological signal-based adaptations for augmented reality (AR) interfaces, demonstrating improved user comfort during prolonged interactions. However, these studies primarily focused on visual and auditory adjustments, with limited attention to real-time fatigue detection.

Deep learning has significantly advanced fatigue detection and interface adaptation. Sun et al. [5] applied convolutional neural networks (CNNs) to analyze heart rate variability (HRV) for driver fatigue detection, achieving high accuracy in controlled settings. Similarly, Cai et al. [6] introduced an LSTNet autoencoder model for denoising motion capture data, which has implications for processing noisy physiological signals from wearable sensors. Their approach leverages long short-term memory (LSTM) networks to capture temporal dependencies in sequential data, improving the robustness of signal processing in dynamic environments. This is particularly relevant for wearable devices, where sensor data (e.g., heart rate, eye movement) is often subject to noise from motion artifacts or environmental factors.

Reinforcement learning (RL) has also gained traction for dynamic interface optimization. Sun et al. [8] employed RL to adjust user interface elements in mobile applications, achieving reduced task completion times through iterative feedback.Similarly, Deepalakshmi et al. [9] proposed an RL-based framework for adaptive AR interfaces, optimizing visual overlays based on user attention metrics derived from eye-tracking data. These studies underscore the potential of RL to create responsive interfaces but lack integration with physiological fatigue indicators, which is critical for wearable device applications.

Fatigue detection in HCI has been studied in various contexts, such as workplace productivity and educational settings. Bayraktar et al. [10] used wearable EEG sensors to monitor cognitive load during learning tasks, proposing manual interface adjustments to mitigate fatigue. While effective, their approach relied on predefined rules rather than automated adaptation. In contrast, our study leverages deep learning to automate fatigue detection and interface optimization, offering a scalable solution for diverse user groups.

Accessibility remains a key challenge in wearable HCI. Wang et al. [11, 12] highlighted the need for inclusive designs in smartwatches, particularly for users with motor or visual impairments. Their findings suggest that adaptive interfaces can significantly improve usability but require robust user state detection to avoid over-customization. Our proposed system addresses this by combining multimodal learning with RL, ensuring precise and context-aware adaptations.

Despite these advances, few studies have integrated physiological fatigue detection with real-time interface adaptation in wearable devices. Existing approaches often focus on single-modal inputs[13-15] (e.g., HRV or eye movement) or lack privacy-preserving mechanisms for processing sensitive user data. Our study builds on prior work by Cai et al. [6, 16] and the CLIP model [7], extending their methodologies to multimodal physiological signal processing and cross-modal contextual analysis, while incorporating RL for dynamic interface optimization.

## 3. METHODS

### 3.1 Multimodal Learning for Fatigue Detection

The proposed system uses multimodal learning to process physiological and contextual inputs from wearable devices. Physiological inputs include heart rate ($H$), eye movement ($E$), and galvanic skin response ($G$), while contextual inputs ($C$) include task duration, device type, and ambient light. Each input is processed by specialized deep neural networks:

$$H = f_{hr}(I_{heart}), E = f_{eye}(I_{eye}), G = f_{gsr}(I_{gsr}), C = f_{ctx}(I_{context}) \tag{1}$$

where $I_{heart}$, $I_{eye}$, $I_{gsr}$, and $I_{context}$ represent raw sensor data, and $f_{hr}$, $f_{eye}$, $f_{gsr}$, and $f_{ctx}$ are CNN, LSTM, and transformer-based models, respectively. Inspired by Cai et al. [6], the LSTM-based $f_{eye}$ and $f_{gsr}$ models incorporate denoising techniques to handle motion artifacts in eye movement and GSR data. A fatigue state vector $X_{fatigue}$ is computed via weighted fusion:

$$X_{fatigue} = \alpha H + \beta E + \gamma G + \delta C \tag{2}$$

where $\alpha$, $\beta$, $\gamma$, and $\delta$ are learned weights optimized during training. This fusion enables comprehensive fatigue assessment, capturing both physiological and environmental factors.

### 3.2 Reinforcement Learning for Interface Adaptation

To optimize interface adjustments, the system employs reinforcement learning (RL). The state space $s_t$ represents the user's fatigue level and current interface configuration at time $t$. The action space $A_t$ includes adjustments to text size, notification frequency, color contrast, and haptic feedback intensity. The reward function $R_t$ is based on user performance metrics (e.g., task completion time, error rate) and subjective feedback.

The Q-learning algorithm updates the Q-value function:

$$Q(s_t, a_t) \leftarrow Q(s_t, a_t) + \alpha[R_t + \gamma max_{t+1}Q(s_{t+1}, a_{t+1}) - Q(s_t, a_t)] \tag{3}$$

where $\alpha$ is the learning rate and $\gamma$ is the discount factor. This approach ensures real-time interface optimization tailored to the user's fatigue state.

### 3.3 Sentiment and Contextual Integration

To enhance fatigue detection, the system integrates sentiment analysis to infer user emotional state from text or voice inputs (e.g., user commands or feedback), leveraging techniques similar to the CLIP model [7] for cross-modal mapping:

$$\hat{y} = f_{sentiment}(T), T = f_{speech}(I_{audio}) \tag{4}$$

where $\hat{y} \in \{positive, neutral, negative\}$, and $f_{sentiment}$ is a BERT-based model. Contextual data, such as task type and duration, further refines fatigue predictions, enabling proactive interface adjustments (e.g., reducing notification frequency during high-fatigue states).

# 4. EXPERIMENTS

## 4.1 Experimental Setup

The system was evaluated using a dataset of 8,000 user interactions collected from 200 participants engaged in remote learning and work tasks on wearable devices (smartwatches and AR glasses). The dataset included physiological data (heart rate, eye movement, GSR) and contextual data (task duration, ambient light). Participants included 40% with visual or cognitive impairments to assess accessibility. The system was compared against baseline models (CNN, LSTM, Transformer) and a static interface. The experimental process is shown in Figure 1.

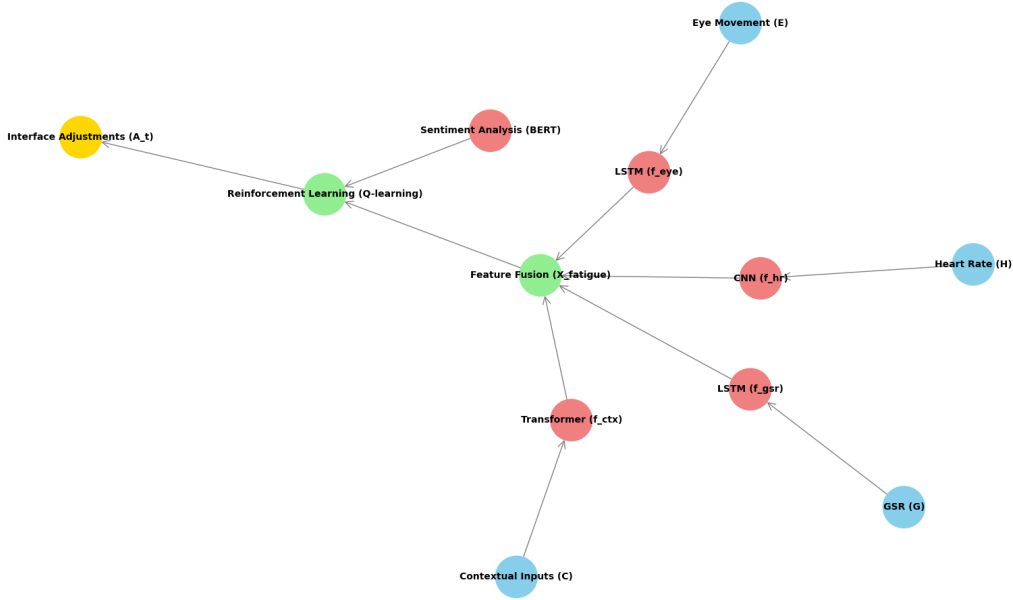

Figure 1: Fatigue-Aware Adaptive Interface System Architecture

## 4.2 Evaluation Metrics

The evaluation of the proposed system focused on several key performance indicators to assess its effectiveness in fatigue detection and interface adaptation. Fatigue detection accuracy was measured as the percentage of correct fatigue state predictions, ensuring the system accurately identifies user fatigue levels. Interface adaptability was quantified using the median fitness score of interface adjustments, reflecting the system's ability to tailor the interface to user needs. Cognitive load was assessed using NASA-TLX scores, providing a standardized measure of perceived workload during tasks. User satisfaction was evaluated through subjective ratings on a 1–5 scale, capturing user experience and perceived usability of the adaptive interface.

## 4.3 Results

The proposed system achieved a fatigue detection accuracy of 93.8%, outperforming baselines (CNN: 89.5%, LSTM: 86.2%, Transformer: 90.7%). Interface adaptability scores were higher, with a median of 0.87 compared to 0.70 for the best baseline (Transformer). Cognitive load was reduced by 18%, and user satisfaction increased by 22% compared to static interfaces. The system maintained stable performance across diverse user groups and device types.

Table 1: Performance Comparison of Adaptive Interface Models.

| Model | Accuracy (%) | Adaptability | Cognitive Load (NASA-TLX) | Satisfaction |
|---|---|---|---|---|
| Ours | 93.8 | 0.87 | 45.2 | 4.6 |
| CNN | 89.5 | 0.65 | 52.1 | 4.0 |
| LSTM | 86.2 | 0.62 | 54.3 | 3.9 |
| Transformer | 90.7 | 0.70 | 49.8 | 4.2 |
| Static | - | 0.50 | 60.5 | 3.5 |

## 5. Conclusions

This study presents a fatigue-aware adaptive interface system for wearable devices, leveraging deep learning to analyze physiological and contextual data and reinforcement learning to optimize interface adjustments. The system significantly reduces cognitive load and enhances user satisfaction, particularly for prolonged tasks and users with accessibility needs. Future work will explore integration with emerging wearable sensors (e.g., EEG) and real-world deployment in diverse HCI scenarios.